# Clustering-Based Approach Extracting Collocations

Mohamed Achraf BEN MOHAMED*, Mounir ZRIGUI** and Mohsen MARAOUI***
* UTIC Laboratory, Monastir, Tunisia. Email: mohamedachraf@wanadoo.fr
** UTIC Laboratory, Monastir, Tunisia. Email: Mounir.zrigui@fsm.rnu.tn
*** UTIC Laboratory, Monastir, Tunisia. Email: maraoui.mohsen@gmail.com

*Abstract*— The following study presents a collocation extraction approach based on clustering technique. This study uses a combination of several classical measures which cover all aspects of a given corpus. It suggests separating bigrams found in the corpus in several disjoint groups according to the probability of presence of collocations. This will allow excluding groups where the presence of collocations is very unlikely and thus reducing in a meaningful way the search space.

## I. INTRODUCTION

Collocation is an expression consisting of two or more words that occur more frequently than by chance. Collocations are important for natural language generation, computational lexicography, parsing, and corpus linguistic research [1]. Collocations include, among others:

- Proper names: "مَكَّة المُكرّمة" (Mecca)
- Verbal expressions: "أَبْصَر النُّور" (He was born)
- Terminologies: "السَّلامُ عَلَيْكُم" (Peace be upon you).

Frequently collocations are not fully compositional, an expression such as "أَصْحَابُ اليَمِين" (Pious) gives an additional meaning when comparing with separated words "أَصْحَابُ" (Those) and "اليَمِين" (on the right).

Several techniques was developed to extracting collocations, however these statistically based measures was penalized by the use of large text samples. In this work we introduce a hybrid clustering based approach allowing to meaningfully reducing the search set.

## II. CONVENTIONAL APPROACHES EXTRACTING COLLOCATIONS

In the following we will introduce some of the accurate approaches to finding collocations.

### A. Hypothesis testing

Hypothesis testing is used to verify whether two words $W_1$ and $W_2$ occur due to association not just by chance. A null hypothesis of independence $H_0$ between $W_1$ and $W_2$ is introduced. All what is needed then is to test the probability of this hypothesis. This task will be ensured by the following statistical measures.

*1) The t test:* The t test is used to compare means of two groups of a normal distribution. In collocations identification, the t test uses to evaluate the difference between the observed (experiment) and expected (given $H_0$) mean. If t is large enough then $W_1$ and $W_2$ are associated. The t test is given by:

$$t = \frac{\overline{X} - \mu}{\sqrt{\frac{S^2}{N}}} \qquad (1)$$

Where $\overline{X}$ is the simple mean, $S^2$ is the sample variance, N is the sample size, and μ is the mean distribution.

*2) Likelihood ratio:* The likelihood ratio is another method for hypothesis testing. The particularity of this measure is its accuracy even with small text samples, thus it is appropriate for identifying occurrence of both common and rare phenomenon [2]. The likelihood ratio allows, being given two hypotheses, to test which one is most likely. In the case of collocations finding the two hypotheses $H_1$ and $H_2$ are [1]:

- $H_1$: independence between $W_1$ and $W_2$:
  $P(w_2|w_1) = P(w_2|\neg w_1) = p$
- $H_2$: dependence between $W_1$ and $W_2$:
  $P(w_2|w_1) = p_1 \neq p_2 = P(w_2|\neg w_1)$

The likelihood ratio is:

$$\lambda = \frac{L(H_1)}{L(H_2)} \qquad (2)$$

Where L is the likelihood function, assuming a binominal distribution L is given by:

$$L(p;n,r) = r^p(1-r)^{n-p} \qquad (3)$$

Where n is the number of trials, r the number of successes, and p is the probability of success.
We can write:

$$\lambda = \frac{L(p;n_1,r_1)L(p;n_2,r_2)}{L(p_1;n_1,r_1)L(p_2;n_2,r_2)} \qquad (4)$$

We use $-2\log\lambda$ instead of $\lambda$ as it's asymptotically $\chi^2$ distributed in the case of binominal distribution [2]. The log likelihood will then have this form:

$$-2\log\lambda = 2[\log L(p_1;n_1,r_1) + \log L(p_2;n_2,r_2) - \log L(p;n_1,r_1) - \log L(p;n_2,r_2)] \quad (5)$$

Where $p_1 = r_1/n_1$, $p_2 = r_2/n_2$, and $p = (r_1+r_2)/(n_1+n_2)$

*B. Mutual information*

Mutual information is another approach allowing identifying collocations. This measure is not based on hypothesis testing, it aims to compare the probability of observing $W_1$ and $W_2$ together ($P(W_1,W_2)$) with the probabilities of observing $W_1$ and $W_2$ independently ($P(W_1)P(W_2)$) [4]. Mutual information is given by:

$$I(W_1,W_2) = \log_2 \frac{P(W_1,W_2)}{P(W_1)P(W_2)} \quad (6)$$

The Mutual information gives how much $W_1$ tell us about $W_2$ [1]. If mutual information is enough large then $W_1$ and $W_2$ are associated else if it is too low then $W_1$ and $W_2$ are independent.

### III. THE PROPOSED APPROACH

The principle of this approach is the combination of several collocation finding measures that complement each other in order to cover all types of collocations that can be found in a corpus. For our experiment we chose the three previously mentioned measures, namely the t test, the likelihood ratio and the mutual information. Then for each collocation candidate $W_1W_2$ these three measures will be carried out to assess the degree of dependence between $W_1$ and $W_2$. From this point we can consider all bigrams extracted from the corpus as being a set of points in a three-dimensional space where each measure represents a dimension and, the problem of collocations identification will be reduced to a clustering problem. More precisely, our work consists in removing the subsets which none of the measures indicates the presence of collocations. Work will be performed in two steps:

- Bigrams extraction and computing of the measures.
- Clustering and exclusion of inappropriate subgroups.

Notations used are summarized in the following:
- *T: Corpus size*
- $L_i$: *lexeme i, $1 \le i \le T$*
- $B_i$: *Bigram i*
- *SL: Stop List*
- $E_i$: *Point i*
- $\mathcal{N}$: *Normalization function*
- $c_k$: *Centroid k*
- $G_k$: *Cluster k*

*1) Bigrams extraction and computing measures*: Being a basic operation, the extraction of bigrams is a delicate process that will directly affect the final result. It first starts by segmenting the corpus, step of identifying the basic units forming the corpus. This means identifying the separators used to isolate morphemes. This step is not exempt from ambiguity [5]. We also adopted a Stop List allowing omitting words you want to ignore because they can not form a collocation as:

- The particles of coordination (ثمّ, أو, أم, أمّا, إمّا...).
- The interrogative particles (أيّ, كيف, أين, متى...).
- The particles of Appeal (يا, أيا, أي, هيا...).
- Prepositions (من, إلى, عن, على, ربّ, حتّى...).
- Conditional particles (مهما, كيفما, حيثما...).

Once the bigrams are identified, the next step is the computation, for each element, of the three measures mentioned previously. These values will be normalized for constraints of graphical representation. Figure 1 summarizes the algorithm.

```
1. //Bigrams extraction and Measures computation
2. for all lexemes l_i, 1≤i≤T-1 do
3.     B_j={ l_i, l_{i+1} / l_i ∉ SL ∧ l_{i+1} ∉ SL}
4. end
5. for all bigrams B_i do
6.     E_i={ N(Mutual_information(B_i)), N(t_test(B_i)),
           N(Likelihood_ratio(B_i))}
7. end
```

Figure 1. Bigrams extraction and Measures computation algorithm

*2) Clustering and exclusion of inappropriate subgroups:* To classify the bigrams we have chosen the Expectation Maximization (EM) algorithm. Once this classification performed, we will browse all subgroups to identify which will have to be excluded. Each group is represented by its centroid. Any group with a centroid located beyond a threshold level will be retained, otherwise it will be excluded. The threshold adopted in our work is to have at least one measure that exceeds 30%. Figure 2 summarizes the algorithm.

```
1. // Clustering
2. G_k=EM_Clustering(data)
3. // Inappropriate subgroups exclusion
4. for all groups G_k do
5.     G_Excluded=G_Excluded U{G_k / c_k < threshold }
6. end
```

Figure 2. Clustering and exclusion of inappropriate subgroups algorithm

### IV. EXPERIMENTS

We have conducted our work on a corpus of 267,131 words covering several topics (economy, sport, religion...). We extracted from this corpus 20,247 bigrams. Clustering operation is schematized by the figure 3. On the same figure we can see excluded clusters (cluster #4, cluster #6, and cluster #8).

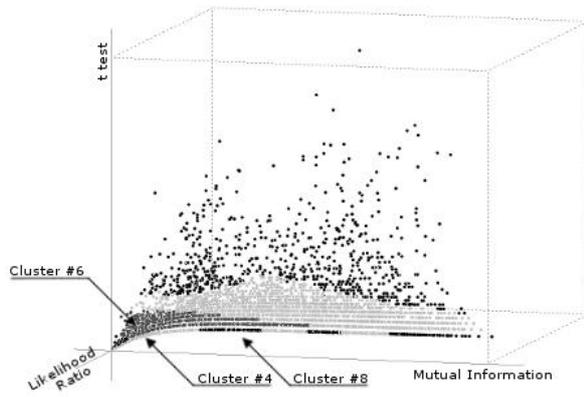

Figure 3.  3D scatter plot for clustered bigrams

It is interesting to note that the dispersed points (bigrams) the highest of Figure 3 represent the elements that the clustering algorithm could not classify and which are considered as noise. This unclassified set contains the largest concentration of collocations. This can be interpreted by the fact that collocations are kind of irregularities of the natural language [3].

Table I summarizes the obtained result. This approach aims to prune incompatible bigrams groups; its use has helped reducing the search space of 34.60%. This technique allows concluding also that the remaining 65.40% are consistent collocations candidates.

TABLE I.
SUMMARY TABLE

| $\Sigma B_i$ | Candidate | Cluster # | $\Sigma|G_k|$ | % |
|---|---|---|---|---|
| 20247 | Yes | 1,2,3,5,7,9 | 13241 | 65.40 |
|  | No | 4,6,8 | 7006 | 34.60 |

TABLE II.
SAMPLE OF CANDIDATE COLLOCATIONS

| $G_k$ | $W_1W_2$ | M.I. | T.T | L.R. |
|---|---|---|---|---|
| 1 | رسول الله | 0,325161 | 0,477212 | 0,818655 |
| 1 | القرآن الكريم | 0,548454 | 0,246556 | 0,388834 |
| 1 | رام الله | 0,224567 | 0,300380 | 0,190636 |
| 2 | السجل الطبي | 0,482222 | 0,145600 | 0,114671 |
| 2 | البحر الميت | 0,549022 | 0,135046 | 0,113905 |
| 3 | قانون المرافعات | 0,556101 | 0,055139 | 0,021825 |
| 3 | رئيس الكورتيس | 0,554995 | 0,055138 | 0,021782 |
| 5 | ستريت جورنال | 0,999999 | 0,055216 | 0,042550 |
| 5 | القيد والعمولة | 0,999999 | 0,055216 | 0,042550 |
| 7 | إسماعيل الأزهري | 0,597662 | 0,055169 | 0,021646 |
| 7 | داخل البناية | 0,593927 | 0,055167 | 0,021499 |
| 9 | أشراط الساعة | 0,731139 | 0,055206 | 0,028753 |
| 9 | قراءة الفنجان | 0,731139 | 0,055206 | 0,028753 |

Table II and Table III respectively give samples of collocation candidates, and excluded bigrams extracted from different clusters.

TABLE III.
SAMPLE OF EXCLUDED BIGRAMS

| $G_k$ | $W_1W_2$ | M.I. | T.T | L.R. |
|---|---|---|---|---|
| 4 | بعض أصحاب | 0,231679 | 0,051646 | 0,006030 |
| 4 | تجاوز سنة | 0,231334 | 0,051631 | 0,006020 |
| 6 | جميع دول | 0,122820 | 0,059803 | 0,004585 |
| 6 | قبل عناصر | 0,126451 | 0,060571 | 0,004778 |
| 8 | داخل مقر | 0,249446 | 0,052322 | 0,006660 |
| 8 | يجب إجراء | 0,249535 | 0,052325 | 0,006659 |

## V. CONCLUSIONS

We realized during this work a method for overcoming a recurrent problem of natural language processing which is handling of large volumes of texts long regarded as an obstacle. The approach adopted opens the way for other methods of collocations extraction, such as the adoption of a POS filter [1], which will further refine the result. The choice of basic measures is also discussed. We chose three measures satisfying an acceptable coverage of a corpus. Indeed, we chose to use the commonly used t test. We improve this choice by the likelihood ratio, sensitive to rare phenomena. Finally we called the mutual information measure which is firstly, not part of the hypothesis testing class; secondly it is sensitive to sparseness [1].

REFERENCES

[1] Manning, Christopher D., and Hinrich Schutze (1999) Foundations of statistical natural language processing. Cambridge, Mass. MIT Press.
[2] Dunning, Ted (1993) Accurate methods for the statistics of surprise and coincidence. Computational Linguistics 19:61-74.
[3] Ellis, N. C., Frey, E., & Jalkanen, I. (2007). The Psycholinguistic Reality of Collocation and Semantic Prosody (1): Lexical Access. In U. Römer & R. Schulze (Eds.), Exploring the Lexis-Grammar interface. Hanover: John Benjamins.
[4] Church, Ken, and Hanks, Patrick (1989). Word association Norms, Mutual Information, and Lexicography. In Proceedings of the 27th Annual Meeting of the Association for Computational Linguistics, p. 76-83, Vancouver, Canada.
[5] Sinclair, J. (1991). Corpus, concordance, collocation. Oxford: Oxford University Press.